\documentclass[runningheads]{llncs}

\usepackage{eccv}

\usepackage{multirow}
\usepackage[table]{xcolor}
\usepackage{arydshln}

\definecolor{pointgreen}{RGB}{34,139,34} 
\definecolor{pointred}{RGB}{178,34,34} 

\usepackage{eccvabbrv}

\usepackage{graphicx}
\usepackage{booktabs}

\usepackage[accsupp]{axessibility}  

\usepackage[pagebackref,breaklinks,colorlinks,citecolor=eccvblue]{hyperref}

\usepackage{orcidlink}

\begin{document}

\title{4D-RaDiff: Latent Point Diffusion for 4D Radar Point Cloud Generation} 

\titlerunning{4D-RaDiff: Latent Point Diffusion for 4D Radar Point Cloud Generation}

\author{Jimmie Kwok\inst{1, 2} \and
Holger Caesar\inst{1}\orcidlink{0000-0001-5099-6297} \and
Andras Palffy\inst{2}\orcidlink{0000-0003-1663-4967}}

\authorrunning{J.~Kwok et al.}

\institute{Delft University of Technology, Netherlands \and
Perciv AI, Netherlands
}

\maketitle

\begin{abstract}
Automotive radar has shown promising developments in environment perception due to its cost-effectiveness and robustness in adverse weather conditions. However, the limited availability of annotated radar data poses a significant challenge for advancing radar-based perception systems. To address this limitation, we propose a novel framework to generate 4D radar point clouds for training and evaluating object detectors. Unlike image-based diffusion, our method is designed to consider the sparsity and unique characteristics of radar point clouds by applying diffusion to a latent point cloud representation. Within this latent space, generation is controlled via conditioning at either the object or scene level. The proposed 4D-RaDiff converts unlabeled bounding boxes into high-quality radar annotations and transforms existing LiDAR point cloud data into realistic radar scenes. Experiments demonstrate that incorporating synthetic radar data of 4D-RaDiff as data augmentation method during training of object detection models consistently improves performance compared to training on real data only. In addition, pre-training on our synthetic radar data achieves competitive detection performance, providing a promising, scalable approach to reduce dependence on costly manual annotations.


\end{abstract}
\section{Introduction}
\label{sec:intro}

Robust perception is crucial for autonomous vehicles to understand their surroundings and make safe driving decisions. Automotive sensors such as cameras and LiDAR have become the predominant choices for perceiving the environment. However, both sensor types show reduced reliability in adverse weather conditions, including rain, fog, or snow. Camera sensors are also negatively affected in low-light environments~\cite{yoneda2019automated}.

In recent years, 4D millimeter-wave radar has emerged as a promising alternative sensor modality. The longer wavelengths of radar allow it to penetrate atmospheric particles, making it suitable for harsh environmental conditions~\cite{sezgin2023safe}. Modern automotive radars provide processed data in the form of point clouds, with the distinctive ability to also measure the relative radial velocity of an object via the Doppler effect. The inclusion of Doppler information has been shown to significantly improve the performance of radar-based object detection~\cite{palffy2020cnn, palffy2022multi}.
 
\begin{figure}[t] 
    \centering
    \includegraphics[width=0.98\linewidth]{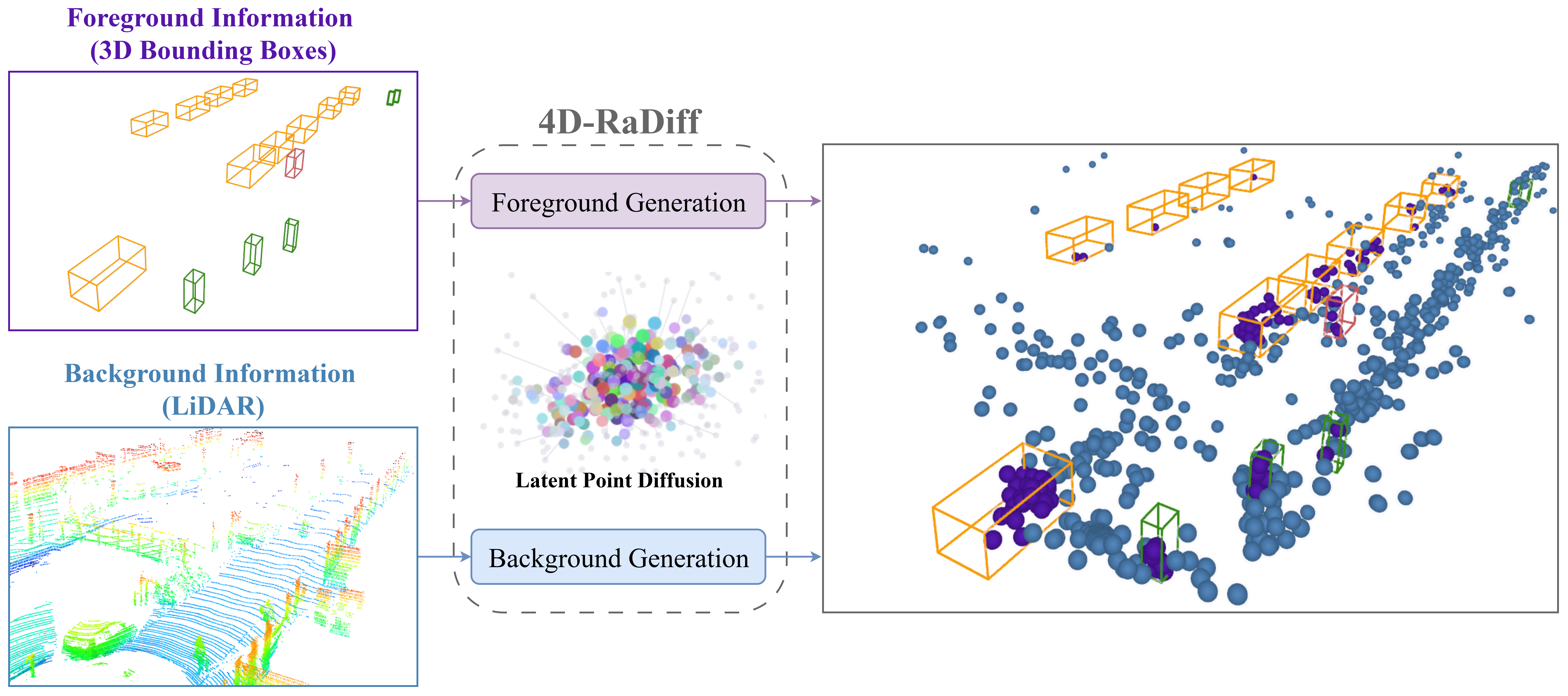}
    \caption{4D-RaDiff separately generates background (\textcolor[RGB]{0,114,189}{blue}) and foreground (\textcolor[RGB]{86,21,170}{purple}) radar points by conditioning on LiDAR and 3D bounding boxes, respectively. Background and foreground points can be fused into a synthetic 4D radar point cloud.} \label{fig:teaser_image}
\end{figure}

Despite the unique attributes of radar, its perception capabilities still fall short of established LiDAR-based methods~\cite{caesar2020nuscenes, palffy2022multi}. This performance gap can be mainly attributed to two key factors: the inherent sparsity of radar point clouds and the limited availability of annotated radar data~\cite{scheiner2021object, zhou2022towards, fent2023radargnn}. While many studies have focused on obtaining denser representations of radar point clouds through generative~\cite{zhang2024towards, wu2024diffradar, luan2024diffusion} or other learning-based approaches~\cite{roldan2024see, han2024denserradar}, they often overlook the scarcity of radar annotations. Acquiring such annotations is labor-intensive, time-consuming, and costly, as it demands extensive effort in data collection, synchronization, and manual labeling. 

To alleviate this problem, we propose 4D-RaDiff, a novel latent point diffusion framework for 4D radar point cloud generation designed for downstream tasks, enabling synthetic data to be readily used for training radar perception models. As shown in Fig.~\ref{fig:teaser_image}, we produce the synthetic data in two ways: (1) synthesizing foreground points within annotated 3D bounding boxes, and (2) generating background points conditioned on LiDAR point clouds to capture scene-level context outside the annotated regions. We disentangle the generation process because foreground and background radar points follow different distributions. Foreground points correspond to semantic objects, often dynamic in nature and composed of distinct reflective surfaces, whereas background points are largely static and dominated by noise and clutter.
Therefore, modeling each component separately enables high-fidelity generation tailored to their underlying characteristics.


For the diffusion process of point cloud data, image representations of point clouds~\cite{004ran2024LiDMs, 001zyrianov2024lidardm, 002hu2024rangeldm, 003nakashima2024lidar, luan2024diffusion} are widely used, as they allow the use of conventional image-based diffusion methods. Nevertheless, radar data produces extremely sparse images, which limits the effectiveness of this approach. A parallel work, RadarGen~\cite{borreda2025radargen}, circumvents this problem by convolving sparse radar point clouds expressed as an image-like bird's-eye view (BEV) representation with a Gaussian kernel to obtain a set of dense image maps, enabling the generation of radar point clouds from multi-view camera images. However, the sparsity of radar point clouds causes RadarGen to spend most of its computation on empty regions, resulting in substantial training and generation times. While applying diffusion directly on the points~\cite{nunes2024scaling} could overcome this problem, modeling the irregular distribution of all radar features remains challenging. Therefore, to model radar point clouds effectively, we operate directly on a point-based latent representation rather than converting them to images. To this end, we first train a point-based variational autoencoder (VAE) to map radar points into a regularized latent space, and subsequently train a diffusion model on these latent representations conditioned on the relevant inputs.

We validate the effectiveness of 4D-RaDiff by evaluating how well synthetic data generated by our approach improves radar-based 3D object detection. We conduct experiments by training jointly on synthetic and real data via data augmentation, and by pre-training on synthetic data followed by fine-tuning on real data.

In summary, our work consists of the following key contributions:
\begin{itemize}
    \item We propose 4D-RaDiff, the first latent point diffusion framework operating directly on a point-based radar representation, allowing for the generation of 4D radar point clouds including Doppler and RCS features.
    \item We introduce a foreground–background generation pipeline, where foreground radar points are generated conditioned on 3D bounding boxes, and background points are synthesized using LiDAR data as conditioning.
    \item We demonstrate that our framework improves radar-based 3D object detection performance on the View-of-Delft~\cite{palffy2022multi} and TruckScenes~\cite{fent2024truckscenes} datasets, using synthetic data for either data augmentation or pre-training.
\end{itemize}
\section{Related Work}
\label{sec:related_work}

\noindent
\textbf{Data Augmentation} artificially inflates the size and diversity of the training dataset, thereby reducing overfitting, improving generalization, and enhancing model performance. In the image domain, augmentation techniques have been extensively studied \cite{shorten2019survey}, whereas data augmentation for point clouds remains relatively underexplored. Basic augmentation methods originally developed for images can also be applied to point cloud data. Common examples are flipping, scaling, rotation, translation, adding random noise, and randomly dropping points. Inspired by mix-based image augmentations \cite{zhang2017mixup, yun2019cutmix, ghiasi2021simple}, GT-Sampling \cite{yan2018second} inserts ground truth objects, sampled from a database constructed prior to training, by copy-pasting them into training samples. The downside of GT-Sampling is that it may still place objects in unreasonable locations, causing either occlusion or collision with the environment. Thus, several improved variants \cite{fang2021lidaraug, hu2023context, lee2023resolving} have been implemented to address these issues. Other augmentations simulate the effects of adverse weather on LiDAR point clouds \cite{kilic2021lidar, hahner2021fog, hahner2022snow}. Despite the existing work on point cloud augmentations, many of the methods were designed for LiDAR point cloud data. For instance, \cite{palffy2022multi} highlights the potential risks of applying simple augmentation strategies to radar point clouds, as they may distort the Doppler feature of the points. Consequently, there is a need for advanced methods that can augment radar point clouds in a way that preserves their semantic features. \\

\noindent
\textbf{Diffusion Models} have garnered significant interest because of their impressive results in image generation~\cite{ho2020denoising, dhariwal2021diffusion, song2021scorebased, nichol2021improved, song2021ddim}. The generation process can be guided by providing conditional information, facilitating tasks such as class-conditional image generation~\cite{dhariwal2021diffusion, karras2022elucidating}, text-to-image generation~\cite{nichol2021glide, saharia2022photorealistic, ramesh2022hierarchical}, and image-to-image translation~\cite{choi2021ilvr, saharia2022palette}. One shortcoming of diffusion models is their high demand for computational resources, particularly when operating in the high-dimensional space of images, which results in prolonged training and inference times. To mitigate this issue, Latent Diffusion Models (LDMs)~\cite{rombach2022high} apply the diffusion process in a lower-dimensional latent space, which significantly improves efficiency while retaining generation quality. \\

\noindent
\textbf{LiDAR-based Diffusion Models} extend the diffusion process to LiDAR point cloud data. The most common data representation of LiDAR points is an image representation, which allows standard image-based diffusion methods to be applied. Both~\cite{003nakashima2024lidar} and~\cite{005helgesen2024fast} focus on LiDAR point cloud upsampling by formulating it as an image inpainting task, where the sparse point cloud serves as the inpainting mask. LidarDM~\cite{001zyrianov2024lidardm}, LiDMs~\cite{004ran2024LiDMs}, and RangeLDM~\cite{002hu2024rangeldm} leverage LDMs to generate LiDAR scenes. LidarDM synthesizes temporally coherent LiDAR sequences conditioned on semantic layout maps. LiDMs are capable of generating LiDAR point clouds from multi-modal conditions (e.g., semantic maps, camera views, and text prompts). RangeLDM addresses the tasks of LiDAR point cloud upsampling and inpainting. Instead of representing LiDAR point clouds as an image,~\cite{nunes2024scaling} applies diffusion directly on the points to generate a more dense LiDAR scene. Other point-based diffusion methods~\cite{kirby2024logen, xiang2025pcd} generate LiDAR point clouds for individual objects rather than entire scenes. In contrast, we operate the diffusion process within a learned latent point cloud representation, which better preserves the unique properties of radar data. \\ 

\noindent
\textbf{Radar-based Diffusion Models} operate on radar data, represented as either a tensor or point cloud. The radar tensor contains power measurements across spatial and Doppler domains and can be interpreted as an image. To obtain the radar point cloud, the Constant False Alarm Rate (CFAR) algorithm is typically applied on this low-level data. Most radar diffusion methods aim to generate denser radar point clouds, using LiDAR as supervision while conditioning on either radar tensor~\cite{zhang2024towards, yang2025unsupervised, wang2025sddiff} or radar point cloud~\cite{wu2024diffradar, luan2024diffusion, wu2025diffusion, zheng2025r2ldm}. Sem-RaDiff~\cite{zhang2025sem} exploits both radar point cloud and tensor information to derive an intermediate representation, which is then used as a conditional input for the diffusion model to generate LiDAR-like point clouds with semantic labels. Unlike these prior works generating a denser representation of radar point clouds, our method generates radar point clouds from scratch, requiring no radar data as conditioning input. RadarGen~\cite{borreda2025radargen} generates radar point clouds from multi-view camera images, but applying image-based diffusion to sparse radar data leads to inefficient use of computation and slow generation speed. In contrast, our point-based latent representation naturally accommodates the sparsity of radar data, enabling both efficient training and fast generation.
\section{Method}
\label{sec:method}

\begin{figure}[t] \centering
    \includegraphics[width=0.98\linewidth,height=0.4509\linewidth]{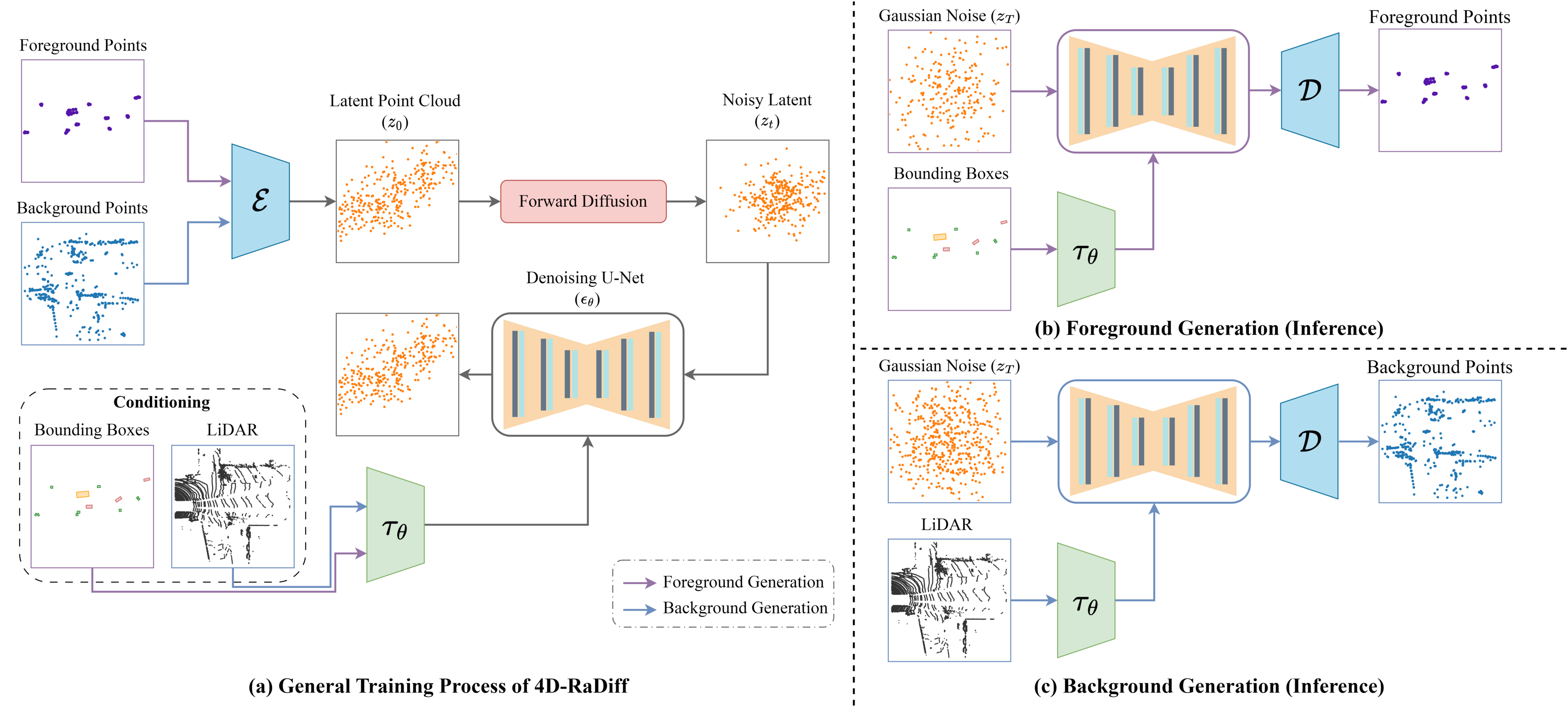}
    \caption{\textbf{The overall framework of 4D-RaDiff.} (a) During the training phase, we encode each radar point cloud with the frozen point-based encoder \(\mathcal{E}\) of the variational autoencoder into a latent point cloud representation. We train two separate latent diffusion models (LDMs) and their respective conditioning encoder \(\tau_\theta\), each tailored to the specific task of either foreground (cf. Sec.~\ref{sec:method_foreground}) or background (cf. Sec.~\ref{sec:method_background}) generation. During inference, the LDM generates foreground points (b) conditioned on 3D bounding boxes, or background points (c) conditioned on the LiDAR point cloud.} \label{fig:model_achitecture}
\end{figure}

In this section, we describe the overall pipeline of the proposed 4D-RaDiff, as shown in Fig.~\ref{fig:model_achitecture}. 

\subsection{Latent Point Diffusion Model}
To address the sparsity of radar point clouds and the highly irregular distribution of their features, existing methods, i.e., image-based diffusion~\cite{004ran2024LiDMs, 001zyrianov2024lidardm, 002hu2024rangeldm, 003nakashima2024lidar, luan2024diffusion} and diffusion in the original point cloud space~\cite{nunes2024scaling, kirby2024logen} may not be well suited to handle these limitations. We circumvent these challenges by applying diffusion that operates within a regularized latent point cloud representation of radar points. To this end, we design a point-based VAE built upon~\cite{he2022density}, and use Sparse Point-Voxel Diffusion (SPVD)~\cite{romanelis2024efficient} as the backbone of our model.

First, we train a VAE to encode the radar point cloud representation into a regularized latent point cloud space. The training objective consists of a reconstruction loss and Kullback-Leibler regularization. The input to the VAE is a radar point cloud, represented as a set of N points with xyz-coordinates, Doppler and RCS as its features, and is denoted as \(x \in \mathbb{R}^{N \times 5}\). The encoder \(\mathcal{E}\) maps the radar point cloud \(x\) to a latent representation \(z = \mathcal{E}(x)\), where \(z \in \mathbb{R}^{M \times d_z}\). The decoder aims to reconstruct the original point cloud from the latent, \(\tilde{x} = \mathcal{D}(z) = \mathcal{D}(\mathcal{E}(x))\). The full training objective and model architecture of the VAE are provided in the supplement.

In the second stage, we train the model \(\epsilon_\theta\) on the latent space representation \(z\), where the VAE's encoder and decoder are kept frozen. We follow the standard formulation of denoising diffusion probabilistic models \cite{ho2020denoising}, with derivations provided in our supplementary material. The objective of our LDM is defined as follows:

\begin{equation}
L_{\text{LDM}} = \mathbb{E}_{\mathcal{E}(x), y,\, \epsilon \sim \mathcal{N}(0, I),\, t} \left[ \left\| \epsilon - \epsilon_\theta(z_t, t, \tau_{\theta}(y)) \right\|_2^2 \right].
\end{equation}

\noindent
Given a condition \(y\), the model \(\epsilon_\theta\) predicts the noise \(\epsilon\) that has been added to the latent point cloud \(z_0\) to produce the noisy latent \(z_t\) after \(t\) forward diffusion steps. We jointly optimize the denoising network \(\epsilon_\theta\) and the encoder \(\tau_\theta\) of the conditioning mechanism, where a separate encoder is designed for each type of conditioning input.

\subsection{Foreground Generation}\label{sec:method_foreground}
Foreground points are defined as the points that lie within the annotated 3D bounding boxes representing road users. Our model generates these points by conditioning on the bounding boxes of the scene. To achieve this, we apply a cross-attention mechanism~\cite{vaswani2017attention} at multiple layers within the backbone of the LDM. 

For the representation and encoder of the bounding boxes, we implement a strategy similar to LayoutDiffusion~\cite{zheng2023layoutdiffusion}, extending their approach from 2D to 3D bounding boxes. Specifically, we define a fixed set of \(n\) objects, denoted as \( \mathcal{S} = \{ o_0, o_1, \ldots, o_{n-1} \} \). Each object is given by \(o_i = \{ b_i, c_i\}\), where \(b_i \in [0, 1]^9 \) is a vector representing the position, size, yaw and 2D velocities of a bounding box and \(c_i \in [0, C+1]\) denotes its class id. We incorporate additional features such as the yaw angle and bounding box velocities, because they provide crucial information for the synthesis of radar point clouds with accurate Doppler values.  Consistent with the design of LayoutDiffusion, the first object \(o_0\) of the set \(\mathcal{S}\) is always a fixed representation with \(c_0 = 0\), where \(b_0\) corresponds to a bounding box covering the entire point cloud range. This fixed representation provides global scene context for conditioning. To ensure a fixed sequence length of \(n\) objects in each frame, we pad the set with empty objects \(o_p\) when necessary, which consists of a zero vector for \(b_p\) and class id \(c_p = C + 1\).

The encoder processes the input set \(\mathcal{S}\) by separately embedding the bounding box and class information for each object and summing them to produce the initial layout embeddings. To encourage the interaction between the different objects, the layout fusion module introduced in~\cite{zheng2023layoutdiffusion} applies multiple layers of self-attention to these embeddings. The embedding corresponding to \(o_0\), which encodes the overall bounding box layout, is used for global conditioning. Subsequently, the embeddings of the objects are fed into the cross-attention layers of the LDM for local conditioning.

\subsection{Background Generation}\label{sec:method_background}
Background points consist of all points in the point cloud that are not classified as foreground points. The generated background points are merged with foreground points to form a complete radar point cloud, which can then be used for downstream tasks. To generate background points, we condition solely on the LiDAR point cloud of the scene. This design choice is motivated by the fact that LiDAR provides a highly accurate representation of the background environment, and most 4D radar point cloud datasets provide both radar and LiDAR modalities. In contrast, alternative sources for background conditioning, such as high-definition maps, are typically not available. Furthermore, this approach enables extending existing datasets that contain only LiDAR data with synthetic radar point clouds.

We employ PointPillars~\cite{lang2019pointpillars} to map the LiDAR point cloud to an intermediate representation, which serves as input to the model’s cross-attention layers. For each LiDAR scan, we consider only points located within the same spatial range as the radar point cloud.
\section{Experiments}
\label{sec:exp}

\textbf{Datasets.} For our experiments, we used the View-of-Delft (VoD)~\cite{palffy2022multi} and TruckScenes~\cite{fent2024truckscenes} dataset. The VoD dataset provides synchronized and calibrated sensor data captured in Delft, which consists of LiDAR point clouds, camera images, and 4D radar point clouds captured from a single front-facing radar sensor. It also contains 3D bounding box annotations, including labels for cars, cyclists, and pedestrians. Although VoD is relatively limited in scale, the TruckScenes dataset offers a much larger and more diverse data collection. The data was primarily recorded on German highways by a truck, obtained using four cameras, six LiDAR, and six radar sensors. It also provides 4D radar point clouds with 360-degree coverage and a broader set of class labels, with an annotation range up to 150 m. \\

\noindent
\textbf{Baselines.} 
While RadarGen~\cite{borreda2025radargen} is the work most closely related to ours, a direct comparison is not straightforward: their code is not publicly available, and they evaluate on a custom data split, making a fair quantitative comparison difficult. Instead, to assess foreground generation quality, we compare against a simple but informed baseline that generates foreground radar points by randomly downsampling LiDAR points within 3D bounding boxes to match radar sparsity, estimating Doppler via radial projection of object velocity, and sampling RCS from class-conditioned statistics. We also propose to evaluate the effectiveness of 4D-RaDiff indirectly, by assessing whether incorporating its synthetic data during training improves performance on a downstream perception task, namely 3D object detection. As baselines, we consider object detectors trained only on real data. We investigate two applications to leverage our synthetic data during training: data augmentation and pre-training solely on synthetic data. For augmentation, we employ GT-Sampling \cite{yan2018second}, a widely used augmentation technique for LiDAR-based object detection. We compare the object detection performance of models trained solely on real data with models trained on both real and synthetic data. The object detection models for these experiments include CenterPoint \cite{yin2021centerpoint}, PointPillars \cite{lang2019pointpillars}, and PillarNet \cite{shi2022pillarnet}. \\  

\noindent
\textbf{Implementation Details.} We train our foreground and background generation models separately on a single RTX 4090 GPU. We train each diffusion model for 1000 epochs using a one-cycle learning rate policy, with a max learning rate of \(1\mathrm{e}{-4}\). For the diffusion process, we use a linear noise schedule with \(T=1000\) diffusion steps, where \(\beta_0 = 1\mathrm{e}{-4}\) and \(\beta_T = 2\mathrm{e}{-2}\). Given the sparsity of radar data, we aggregate multiple radar sweeps for all experiments, following \cite{palffy2022multi} and \cite{fent2024truckscenes}. Furthermore, all object detectors are trained with standard global augmentations by default, including random horizontal flip, global rotation within [\(-\pi/4\), \(\pi/4\)], and global scaling in the range of [0.95, 1.05]. Additional details about the implementation can be found in Appendix~\ref{sup:implementation_details}. \\

\noindent
\textbf{Evaluation Metrics.}
To assess the generative performance of our method, we use Chamfer Distance (CD) to measure the spatial similarity between generated and real radar point clouds. Moreover, we introduce radar-specific metrics based on the Chamfer Distance, \(\text{CD}\) (Doppler) and \(\text{CD}\) (RCS), which quantify the fidelity of the generated Doppler and RCS values, respectively. Please refer to the supplementary material for detailed definitions. For evaluating object detection performance, the VoD dataset uses average precision (AP) with IoU thresholds of 0.5 for cars, and 0.25 for cyclists and pedestrians. VoD also reports metrics for two regions: (1) entire annotated area (camera FoV up to 50 meters) and (2) driving corridor, a safety-critical region located in front of the ego vehicle defined as \([-4\text{ m} < x < 4\text{ m}, 0 < z < 25\text{ m}]\) in camera coordinates. Detection performance on the TruckScenes dataset is measured by a distance-based AP for each class and the nuScenes Detection Score (NDS) \cite{caesar2020nuscenes}.

\subsection{Radar Point Cloud Generation}
In Tab.~\ref{tab:cd_metrics}, we compare 4D-RaDiff to our LiDAR downsampling baseline regarding the generation quality of radar points of objects specifically. While our method achieves better geometric fidelity, it particularly excels in capturing the radar-specific attributes, resulting in substantially higher fidelity in Doppler and RCS across all object classes.

\begin{figure}[t] \centering
\includegraphics[width=0.98\linewidth,height=0.76783\linewidth]{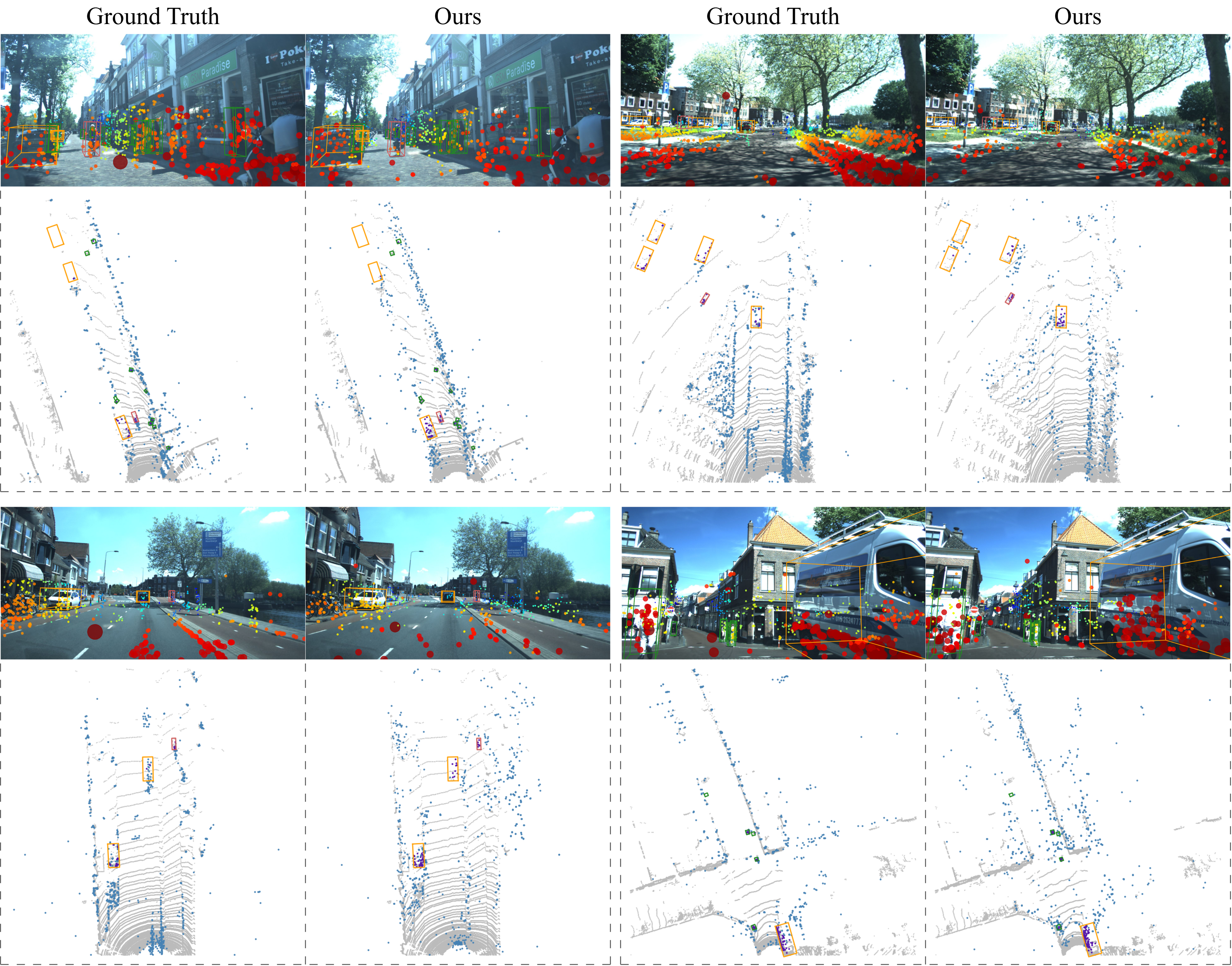}
    \caption{Qualitative results of 4D radar point cloud generation on the VoD dataset. Each camera image shows the radar point clouds projected onto it, while the corresponding bottom image represents the radar point clouds in a bird's-eye view (BEV) of the same scene. In the BEV image, radar point clouds are distinguished as foreground (\textcolor[RGB]{86,21,170}{purple}) and background (\textcolor[RGB]{70, 130, 180}{blue}). 3D bounding boxes are displayed in \textcolor[RGB]{255, 158, 0}{orange} for cars, \textcolor[RGB]{205, 92, 92}{red} for cyclists, and \textcolor[RGB]{34, 139, 34}{green} for pedestrians. LiDAR point clouds are included in the BEV images for reference only.} \label{fig:qualitative}
\end{figure}

\begin{figure}[t] \centering
\includegraphics[width=0.48\textwidth,height=0.554\textwidth]{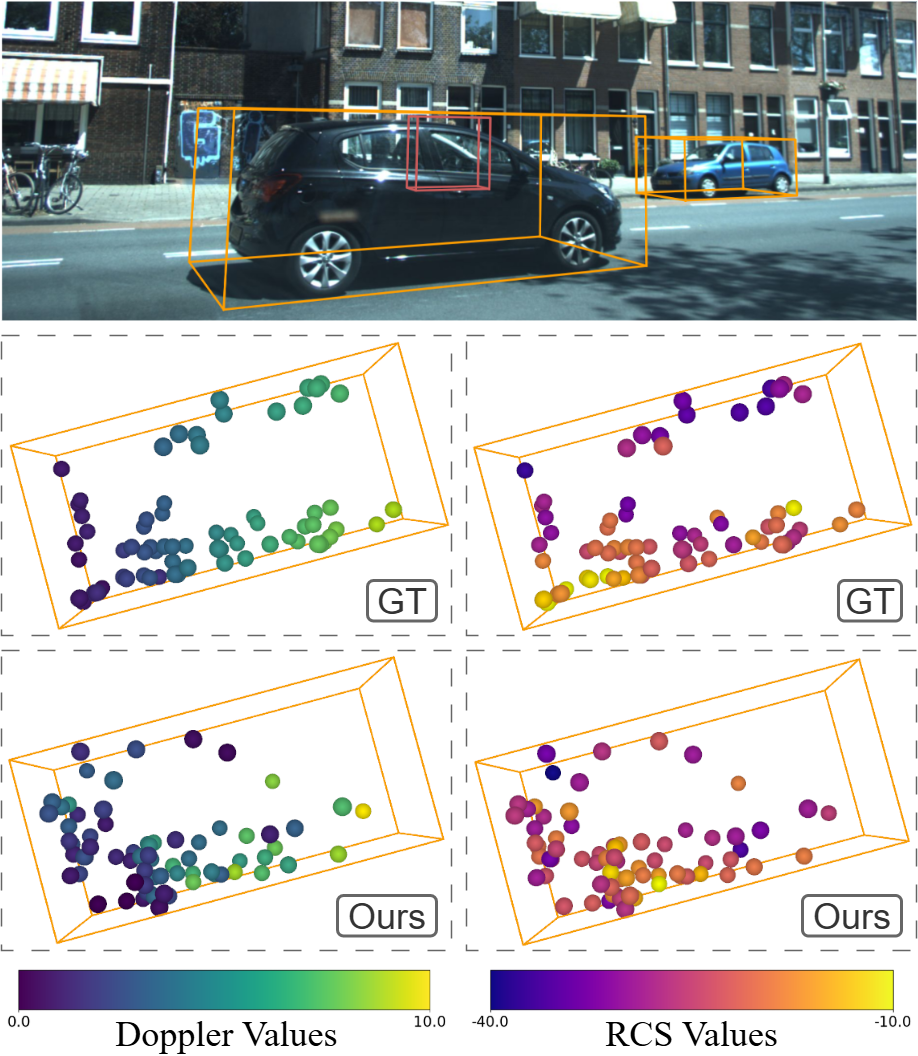}
    \caption{Qualitative example of generated radar points for an object from the VoD validation set. Both real and synthetic points of the black car in front of the ego vehicle are plotted in top view, colored by either Doppler or RCS.} 
    \label{fig:qualitative_foreground}
\end{figure}

In Fig.~\ref{fig:qualitative}, we qualitatively compare our generated radar point clouds against the ground truth, with foreground and background points visualized in distinct colors. We also show a qualitative example of our generated foreground points for an object in the VoD validation set in Fig.~\ref{fig:qualitative_foreground}. Additional qualitative examples of synthetic radar points for various objects from the TruckScenes dataset can be found in Appendix~\ref{sup:add_results}.

\begin{table}[t]
\centering
\caption{Comparison of 4D-RaDiff to LiDAR downsampling baseline for foreground generation on VoD. Evaluation metrics for position (Pos.), Doppler, and RCS fidelity.}
\label{tab:cd_metrics}
\setlength{\tabcolsep}{2.5pt}
\resizebox{0.7\linewidth}{!}{
\begin{tabular}{l c c c c c c c c c}
\toprule
\multirow{2}{*}{\textbf{Method}}
& \multicolumn{3}{c}{\textbf{CD Car} ($\downarrow$)}
& \multicolumn{3}{c}{\textbf{CD Bike} ($\downarrow$)}
& \multicolumn{3}{c}{\textbf{CD Ped.} ($\downarrow$)} \\
\cmidrule(lr){2-4} \cmidrule(lr){5-7} \cmidrule(lr){8-10}
& Pos. & Doppler & RCS
& Pos. & Doppler & RCS
& Pos. & Doppler & RCS \\
\midrule
Baseline
& 0.63 & 4.98 & 17.13
& 0.18 & 1.87 & 10.61
& 0.14 & 0.86 & 8.99 \\
Ours
& \textbf{0.52} & \textbf{2.06} & \textbf{5.69}
& \textbf{0.15} & \textbf{1.28} & \textbf{4.58}
& \textbf{0.11} & \textbf{0.85} & \textbf{4.52} \\
\bottomrule
\end{tabular}
}
\end{table}

\subsection{Augmenting Real Data with Synthetic Data}
In this experiment, we evaluate the quality of our generated foreground points by measuring the performance of an object detector when using these points for augmentation during training. To achieve this, we generate foreground points for the car, cyclist, and pedestrian classes, ensuring a consistent approach across datasets. We use GT-Sampling \cite{yan2018second} to insert synthetic radar points and their corresponding bounding boxes into training samples. A comparable experiment with synthetic data using GT-Sampling has been explored in \cite{jung2025l2rdas}, but our method applies the augmentation strategy to radar point clouds, in contrast to radar tensor data. GT-sampling creates a database prior to training, storing foreground points and bounding boxes from the training dataset. During training, points and their associated bounding boxes are randomly sampled from the database and inserted at their original positions into the point clouds of different scenes. A collision check is performed after insertion to remove any sampled boxes and points that overlap with existing bounding boxes. 

\begin{table}[t!]
\centering
\caption{
3D object detection performance on the VoD validation set. Metrics are reported for the entire annotated area and the driving corridor. For the real and synthetic setting, the database of each contains the same amount of bounding boxes.
}
\label{tab:vod_results}
\resizebox{1\linewidth}{!}{
\begin{tabular}{c lcccccccc}
\toprule
\multirow{2}{*}{\textbf{Category}}
& \multirow{2}{*}{\textbf{Method}}
& \multicolumn{4}{c}{\textbf{Entire Annotated Area}}
& \multicolumn{4}{c}{\textbf{Driving Corridor}} \\
\cmidrule(lr){3-6} \cmidrule(lr){7-10}
& 
& mAP & AP$_{\text{car}}$ & AP$_{\text{cyc}}$ & AP$_{\text{ped}}$
& mAP & AP$_{\text{car}}$ & AP$_{\text{cyc}}$ & AP$_{\text{ped}}$ \\
\midrule
\multirow{2}{*}{\shortstack{Baseline \\ (Real)}}
& CenterPoint \cite{yin2021centerpoint}
& 46.0 & 37.9 & 65.2 & 34.9
& 66.2 & 69.2 & 86.7 & 42.8 \\
& CenterPoint + GT-Sampling \cite{yan2018second} (real)
& 52.4 & 39.9 & 74.1 & 43.2
& 71.0 & 71.2 & 91.1 & 50.8 \\
\midrule
\multirow{2}{*}{\shortstack{Ours \\ (Synthetic)}}
& CenterPoint + GT-Sampling (synthetic)
& 51.2 & 39.8 & 71.5 & 42.3
& 69.8 & 72.0 & 89.5 & 47.8 \\
& CenterPoint + GT-Sampling (real + synthetic)
& \textbf{53.3} & \textbf{40.5} & \textbf{75.0} & \textbf{44.4}
& \textbf{72.5} & \textbf{72.3} & \textbf{93.9} & \textbf{51.2} \\
\bottomrule
\end{tabular}
}
\end{table}

\begin{table}[t!]
\centering
\caption{3D object detection performance on the TruckScenes validation set. The detection range is evaluated up to 75 m for all classes (note that the official range is 150 m for Car, Truck, Trailer, O.V., and Bus). $\dagger$ indicates classes where GT-Sampling is applied in both real (R) and synthetic (S) settings. O.V., Ped., Motor., T.C., T.S. are short for other vehicle, pedestrian, motorcycle, traffic cone and traffic sign, respectively. The official TruckScenes evaluation classes Bus, Barrier, and Animal achieve zero AP for all methods and are omitted from this table, but are included in mAP and NDS calculation.}
\resizebox{1\linewidth}{!}{
\begin{tabular}{l c c c c c c c c c c c c}
\toprule
Method & GT-Sampling \cite{yan2018second} & mAP & NDS & Car$^\dagger$ & Truck & Trailer & O.V. & Ped.$^\dagger$ & Motor. & Bike$^\dagger$ & T.C. & T.S. \\
\midrule

\multirow{2}{*}{CenterPoint \cite{yin2021centerpoint}} 
& R & 19.8 & 27.3 & 55.2 & 45.5 & \textbf{56.5} & 3.5 & 1.9 & 36.3 & 0.0 & 16.9 & \textbf{22.0} \\
& R+S & \textbf{20.8} & \textbf{27.9} & \textbf{55.5} & \textbf{46.0} & 56.3 & \textbf{4.6} & \textbf{4.9} & \textbf{37.5} & \textbf{4.8} & \textbf{18.0} & 21.7 \\
\noalign{\vskip 2.2pt}
\rowcolor{gray!10}
\itshape
Improvements &  & 
\textcolor{pointgreen}{+1.0} & \textcolor{pointgreen}{+0.6} & \textcolor{pointgreen}{+0.3} &
\textcolor{pointgreen}{+0.5} & \textcolor{pointred}{-0.2} &
\textcolor{pointgreen}{+1.1} & \textcolor{pointgreen}{+3.0} &
\textcolor{pointgreen}{+1.2} & \textcolor{pointgreen}{+4.8} &
\textcolor{pointgreen}{+1.1} & \textcolor{pointred}{-0.3} \\
\midrule

\multirow{2}{*}{PointPillars \cite{lang2019pointpillars}}
& R  & 14.7 & 28.4 & 48.3 & 30.5 & 44.5 & 1.6 & 0.0 & \textbf{26.0} & 0.0 & 9.9 & 15.5  \\
& R+S & \textbf{15.3} & \textbf{29.9} & \textbf{49.7} & \textbf{30.8} & \textbf{45.0} & \textbf{2.1} & \textbf{2.3} & 24.9 & \textbf{2.6} & \textbf{10.4} & \textbf{15.8}  \\
\noalign{\vskip 2.2pt}
\rowcolor{gray!10}
\itshape
Improvements &  & 
\textcolor{pointgreen}{+0.6} & \textcolor{pointgreen}{+1.5} & \textcolor{pointgreen}{+1.4} &
\textcolor{pointgreen}{+0.3} & \textcolor{pointgreen}{+0.5} &
\textcolor{pointgreen}{+0.5} & \textcolor{pointgreen}{+2.3} &
\textcolor{pointred}{-1.1} & \textcolor{pointgreen}{+2.6} &
\textcolor{pointgreen}{+0.5} & \textcolor{pointgreen}{+0.3} \\
\midrule

\multirow{2}{*}{PillarNet \cite{shi2022pillarnet}} 
& R  & 20.3 & 25.3 & 55.1 & \textbf{46.6} & \textbf{54.5} & 4.1 & 2.7 & 35.4 & 0.4 & \textbf{18.3} & 26.3 \\
& R+S  & \textbf{21.0} & \textbf{27.8} & \textbf{55.3} & 46.3 & 53.7 & \textbf{4.2} &\textbf{ 6.2} & \textbf{37.0} & \textbf{4.7} & 18.1 & \textbf{26.8} \\
\noalign{\vskip 2.2pt}
\rowcolor{gray!10}
\itshape
Improvements &  & 
\textcolor{pointgreen}{+0.7} & \textcolor{pointgreen}{+2.5} & \textcolor{pointgreen}{+0.2} &
\textcolor{pointred}{-0.3} & \textcolor{pointred}{-0.8} &
\textcolor{pointgreen}{+0.1} & \textcolor{pointgreen}{+3.5} &
\textcolor{pointgreen}{+1.6} & \textcolor{pointgreen}{+4.3} &
\textcolor{pointred}{-0.2} & \textcolor{pointgreen}{+0.5} \\
\bottomrule
\end{tabular}
}
\label{tab:truckscenes_results}
\end{table}

Tab.~\ref{tab:vod_results} reports the object detection performance of CenterPoint \cite{yin2021centerpoint} on the VoD dataset, comparing the effects of GT-Sampling using only real data, only synthetic data, or a combination of both. Using synthetic data alone for GT-Sampling results in slightly lower performance compared to sampling from real data, indicating that our generated object-level radar points closely resemble real points in terms of their effectiveness for training. The best performance is achieved when both real and synthetic data are used jointly as augmentation strategy, demonstrating the benefit of our generative model for improving existing augmentation methods.

To showcase the versatility of our method, we conduct a similar study on the TruckScenes dataset involving multiple object detectors such as CenterPoint \cite{yin2021centerpoint}, PointPillars \cite{lang2019pointpillars}, and PillarNet \cite{shi2022pillarnet}. TruckScenes contains a larger variety of class labels than VoD, but also suffers from class imbalance, with cars accounting for 46.09\%, pedestrians 1.50\%, and bicycles only 0.16\% of all training labels. To mitigate this, we generate foreground points for all three classes, with particular emphasis on oversampling the underrepresented pedestrian and bicycle categories. As shown in Tab. \ref{tab:truckscenes_results}, incorporating our synthetic data leads to significant improvements for the rare classes (pedestrians and bicycles), with minor improvements for the car class. Although some classes may experience slight performance drops when including our generated radar points, overall metrics (mAP and NDS) consistently improve across all object detectors.

\subsection{Pre-Training on Synthetic Data}
\begin{table}[t!]
\centering
\caption{
Comparison of 3D object detection results on the VoD validation set using CenterPoint \cite{yin2021centerpoint} under different training strategies. Metrics are reported for the entire annotated area and the driving corridor. Percentage (\%) indicates the proportion of samples relative to the full VoD training set (5138 samples).}
\label{tab:vod_pretraining}
\resizebox{1\linewidth}{!}{
\begin{tabular}{c c c c c c c c c c c}
\toprule
\multirow{2}{*}{\textbf{Category}}  & \multicolumn{2}{c}{\textbf{Method}} 
& \multicolumn{4}{c}{\textbf{Entire Annotated Area}} 
& \multicolumn{4}{c}{\textbf{Driving Corridor}} \\
\cmidrule(lr){2-3} \cmidrule(lr){4-7} \cmidrule(lr){8-11}
& Pre-train (synthetic) & Fine-tune (real) 
& mAP & AP$_{\text{car}}$ & AP$_{\text{cyc}}$ & AP$_{\text{ped}}$ 
& mAP & AP$_{\text{car}}$ & AP$_{\text{cyc}}$ & AP$_{\text{ped}}$ \\
\midrule
Baseline & 0\%  & 100\%  
& 46.0 & 37.9 & 65.2 & 34.9 
& 66.2 & 69.2 & 86.7 & 42.8 \\
\midrule
\multirow{3}{*}{\shortstack{Synthetic \\ Only}}
& 100\% & 0\% 
& 10.8 & 11.2 & 10.9 & 10.4 
& 14.6 & 25.5 & 6.4 & 12.0 \\
& 200\% & 0\% 
& 13.6 & 14.3 & 13.5 & 12.9 
& 17.2 & 28.5 & 8.2 & 14.9 \\
& 300\% & 0\% 
& 11.0 & 11.4 & 11.4 & 10.2 
& 14.2 & 24.4 & 5.6 & 12.5 \\
\midrule
\multirow{3}{*}{\shortstack{Synthetic + \\ Real (10\%)}}
& 100\% & 10\% 
& 47.0 & 37.5 & 66.9 & 36.7 
& 65.9 & 69.7 & 85.2 & 42.9 \\
& 200\% & 10\% 
& 47.6 & 38.2 & 67.8 & 36.9 
& 66.9 & 70.0 & 86.1 & 44.6 \\
& 300\% & 10\% 
& 46.5 & 36.8 & 67.3 & 35.4 
& 64.9 & 70.0 & 81.3 & 43.4 \\
\midrule
\multirow{3}{*}{\shortstack{Synthetic + \\ Real (100\%)}}
& 100\% & 100\% 
& 47.6 & 37.5 & \textbf{69.4 }& 35.9 
& 67.0 & 70.2 & \textbf{87.2} & 43.7 \\
& 200\% & 100\% 
& \textbf{48.9} & \textbf{39.6} & 69.0 & \textbf{38.2}
& \textbf{67.5} & \textbf{70.7} & 86.1 & \textbf{45.7} \\
& 300\% & 100\% 
& 47.9 & 38.6 & 68.9 & 36.2 
& 66.6 & 70.0 & 84.9 & 44.8 \\
\bottomrule
\end{tabular}
}
\end{table}

We evaluate the quality of both our generated background and foreground points on the VoD dataset by pre-training an object detector on fully synthetic radar point clouds. In Fig.~\ref{fig:qualitative}, we showcase examples of synthetic radar point clouds, which combine the generated background and foreground points of 4D-RaDiff. To obtain our synthetic training data, we use LiDAR from the test split (which is not used for official evaluation) combined with global transformations to generate background points, and we sample random bounding boxes to generate foreground points. 

The results of pre-training a CenterPoint \cite{yin2021centerpoint} detector on fully synthetic radar data are presented in Tab. \ref{tab:vod_pretraining}. We experiment with different amounts of pre-training data, using the same number of synthetic samples as the actual training set (100\%), as well as twice (200\%) and three times (300\%) that amount. We also investigate several fine-tuning configurations: no fine-tuning, fine-tuning on 10\%, and fine-tuning on 100\% of annotated training data. For the 10\%-configuration, we select every tenth sample from the training data, effectively simulating a 1 Hz annotation frequency (given VoD's annotation frequency of 10 Hz).

When training exclusively on synthetic data, we observe a significant performance drop compared to training on real data, emphasizing the domain gap between synthetic and real data. As expected, using twice as much  synthetic data, the performance of the detector increases. Conversely, when training on even more synthetic data (300\%) the performance does not improve.  This suggests that simply increasing the amount of synthetic training data does not always improve performance, as it becomes harder for the model to generalize to real-world data distributions. 

Notably, pre-training on synthetic data followed by fine-tuning on just 10\% of real data results in substantial performance improvements compared to training solely on synthetic data. More importantly, across all pre-training configurations, the model consistently outperforms the baseline trained solely on the full real training set without any pre-training. This indicates that pre-training on synthetic data can initialize the model weights  effectively, enabling it to achieve good performance even in the absence of large annotated data.

Finally, pre-training and fine-tuning on the entire training set achieves the best overall performance. However, the performance gain of using the whole training set (100\%) for fine-tuning compared to fine-tuning on a small subset (10\%) is not as substantial as one might expect. A plausible explanation is that bridging the synthetic-to-real domain gap requires only a small amount of diverse, real annotated data during fine-tuning, as the model has already learned the fundamental structure of the task during pre-training. 

\subsection{Ablation Study and Efficiency Analysis}\label{subsec:ablation}
In this ablation study, we verify the effectiveness of departing to the latent space for the diffusion process of radar point clouds. We evaluate the quality of the generated background radar points on the VoD validation set in Tab.~\ref{tab:ablation_latent}, comparing diffusion applied on radar point clouds directly with our latent diffusion approach. We observe that our approach significantly outperforms the vanilla diffusion model across all metrics, showing the benefit of operating the diffusion within the latent space. We also provide a qualitative comparison of both approaches in the supplementary materials. In addition, an ablation on the backbone of our diffusion model can be found in Appendix~\ref{sup:ablation}.

\begin{table}[t!]\centering
\caption{Ablation on the latent point cloud space as input to the latent diffusion model (LDM) by removing the VAE component. Comparison of background generation of radar point clouds evaluated on validation set from VoD.}
\label{tab:ablation_latent}
\resizebox{0.40\linewidth}{!}{
\begin{tabular}{l c c c c}
\toprule
\multirow{2}{*}{\textbf{Method}}
& \multirow{2}{*}{\textbf{Latent}}
& \multicolumn{3}{c}{\textbf{CD} ($\downarrow$)} \\
\cmidrule(lr){3-5} 
& & Pos. & Dop. & RCS \\
\midrule
SPVD \cite{romanelis2024efficient}
& $\times$ & 3.63 & 4.80 & 25.04 \\
Ours
& \checkmark & \textbf{2.26} & \textbf{1.24} & \textbf{6.75} \\
\bottomrule
\end{tabular}
}
\end{table}

\begin{table}[t!]
\centering
\caption{Training time, throughput (\(T=1000\) sampling steps), and inference speed for different model components and datasets. Throughput is defined as the number of generated samples per second by the model. Inference Speed refers to the number of diffusion steps performed per second by the model. Results are based on using a single RTX 4090 GPU.}
\label{tab:model_time}
\setlength{\tabcolsep}{4pt} 
\begin{tabular}{lccccc} 
\toprule
Model & Task & Dataset & \shortstack{Training Time \\ (h)} & \shortstack{Throughput \\ (samples/s)} & \shortstack{Inference Speed \\ (steps/s)} \\ 
\midrule
VAE & Foreground & VoD & 3.75 & - & - \\
LDM & Foreground & VoD & 13.02 & 1.07 & 1069.3 \\
VAE & Background & VoD & 1.67 & - & - \\
LDM & Background & VoD & 29.12 & 0.25 & 253.5 \\
VAE & Foreground & TruckScenes & 6.83 & - & - \\
LDM & Foreground & TruckScenes & 18.75 & 0.85 & 853.4 \\
\bottomrule
\end{tabular}
\end{table}

Efficient and scalable generation is essential for producing large amounts of synthetic radar data. Tab.~\ref{tab:model_time} provides an overview of the training time, throughput, and inference speed of 4D-RaDiff and its individual components across datasets. For reference, RadarGen~\cite{borreda2025radargen} reports a training time on the TruckScenes dataset of 2 days on 8 L40 (48GB) GPUs with an inference speed of approximately 0.095 steps/s using a single L40. Even though direct comparison is difficult due to differences in hardware, dataset splits, and experimental setup, these numbers highlight the efficiency gains of our latent point diffusion approach.

\subsection{Limitations}
Despite the impressive results in 4D radar point cloud generation, our method still has several limitations. First, our model generates a fixed number of points, which can lead to discrepancies in density between real and generated radar point clouds, as can be seen in the top right of Fig.~\ref{fig:qualitative}. Second, although separating the generation process for background and foreground points helps maintain high fidelity for each component, this detachment can introduce inconsistencies. For example, partial occlusion caused by the environment is not taken into consideration (see Appendix~\ref{sup:add_results}). Third, our foreground generation does not model the trail of motion produced by dynamic objects when aggregating multiple radar scans. This could be addressed by also compensating the motion of dynamic objects based on Doppler information, as done in \cite{haitman2025doppdrive}. Finally, the background generation is conditioned on LiDAR point cloud data. This imposes various practical constraints, as it requires a high-cost sensor and makes our approach sensor-dependent, limiting generalization across datasets. Leveraging alternative conditioning inputs for the background environment, such as semantic maps or camera images, could be a promising direction.
\section{Conclusion}
\label{sec:conclusion}

In this paper, we have presented a novel latent point diffusion framework for generating 4D radar point clouds. The proposed 4D-RaDiff performs diffusion on a latent point cloud representation, conditioned on either 3D bounding boxes or LiDAR point clouds to synthesize corresponding radar points for foreground and background, respectively. Thus, our method can generate radar annotations to support downstream task training. Our experiments demonstrate that synthetic data generated by 4D-RaDiff improves object detection through data augmentation. We also conducted experiments with pre-training strategies on our synthetic data, showing that fine-tuning on minimal real radar data achieves competitive performance. This is particularly valuable given the difficulty and cost of obtaining accurate radar annotations. Overall, 4D-RaDiff provides a scalable solution for acquiring high-quality and diverse training data for radar-based perception, reducing the reliance on extensive manual annotation in autonomous driving. 


%
%
\bibliographystyle{splncs04}
\bibliography{main} 
\clearpage
\clearpage

\section*{Supplementary of ``4D-RaDiff: Latent Point Diffusion for 4D Radar Point Cloud Generation''}

\appendix
\renewcommand{\thesection}{\Alph{section}}
\renewcommand{\thesubsection}{\thesection.\arabic{subsection}}
\renewcommand{\theHsection}{appendix.\Alph{section}}
\renewcommand{\theHsubsection}{appendix.\Alph{section}.\arabic{subsection}}

\noindent
This supplementary material is organized as follows:
\begin{itemize}
    \item In Section~\ref{sup:diff_details}, we provide additional background information on diffusion models.
    \item In Section~\ref{sup:vae}, we describe the architecture and implementation of the variational autoencoder.
    \item In Section~\ref{sup:implementation_details}, we outline further implementation details relevant to our experiments.
    \item In Section~\ref{sup:ablation}, we perform an ablation study on the backbone of our latent diffusion model.
    \item In Section~\ref{sup:add_results}, we present additional qualitative results.
\end{itemize}

\section{Diffusion Model Details}\label{sup:diff_details}
Given a sample \(x_0 \sim q(x)\) from a target data distribution, diffusion models \cite{ho2020denoising} apply a fixed forward diffusion process, formulated as a Markov chain:

\begin{equation}
q(x_{1:T} | x_0) = \prod_{t=1}^{T} q(x_t | x_{t-1}), \quad
\end{equation}

\begin{equation}
    q(x_t | x_{t-1}) = \mathcal{N}\big(x_t; \sqrt{1 - \beta_t}\, x_{t-1}, \, \beta_t I\big),
\end{equation}

\noindent
where noise is gradually added to the input for \(T\) number of steps according to a variance schedule \(\beta_t\). After \(T\) forward steps, the sample will approximately converge to Gaussian noise: \(q(x_T) \approx \mathcal{N}(0, I)\). The forward diffusion process can be simplified using the reparameterization trick \cite{kingma2013auto}: 

\begin{equation}
    q(x_t | x_0) = \mathcal{N}(x_t; \sqrt{\bar{\alpha}_t}\, x_0, \, (1 - \bar{\alpha}_t) I ),
\end{equation}

\noindent
where \(\alpha_t = 1 - \beta_t \) and \(\bar{\alpha}_t = \prod_{s=1}^{t} \alpha_s\). This formulation allows sampling \(x_t\) at an arbitrary timestep \(t\) without computing all intermediate steps.

Diffusion models learn to reverse the forward diffusion:

\begin{equation}
p_\theta(x_{0:T}) = p(x_T) \prod_{t=1}^{T} p_\theta(x_{t-1} | x_t), \quad 
\end{equation}

\begin{equation}
p_\theta(x_{t-1} | x_t) = \mathcal{N}(x_{t-1}; \mu_\theta(x_t, t), \, \beta_tI),
\end{equation}

\noindent
where the mean \(\mu_\theta(x_t, t)\) is defined by the following parameterization:

\begin{equation}
\mu_\theta(x_t, t) = \frac{1}{\sqrt{\alpha_t}} ( x_t - \frac{\beta_t}{\sqrt{1 - \bar{\alpha}_t}} \, \epsilon_{\theta}(x_t, t)).
\end{equation}

\noindent
In the equation above, the model \(\epsilon_{\theta}\) is trained for all possible timesteps \(t\) to predict the noise \(\epsilon\) that has been added to a noisy sample \(x_t\). By doing so, the diffusion model can iteratively denoise a sample after training:

\begin{equation}
x_{t-1} = \frac{1}{\sqrt{\alpha_t}} ( x_t - \frac{\beta_t}{\sqrt{1-\bar{\alpha}_t}} \, \epsilon_\theta(x_t, t)) + \sqrt{\beta_t}\, \mathcal{N}(0, I).
\end{equation}

\noindent
Thus, starting the reverse diffusion from a random noise sample \(x_T \sim \mathcal{N}(0, I)\), the diffusion model iteratively denoises it until a clean generated sample \(x_0\) representative of the target data distribution is obtained.

\section{Variational Autoencoder Implementation}\label{sup:vae}

\subsection{Model Architecture}
\begin{figure*}[tb] \centering
\includegraphics[width=0.98\linewidth, height=0.31615\linewidth]{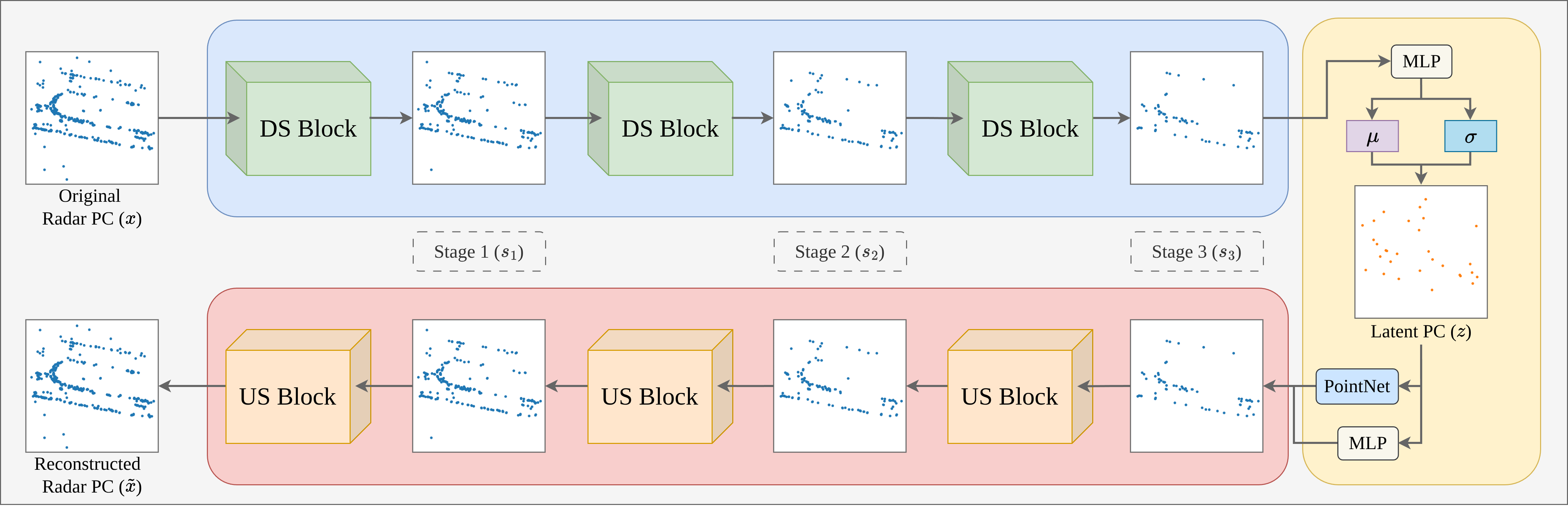}
    \caption{Model architecture of the VAE used in 4D-RaDiff. Downsampling Block (DS) and Upsampling Block (US) are adapted from \cite{he2022density}; please refer to the original paper for details.} 
\label{fig:vae_architecture}
\end{figure*}

The implementation of our VAE builds upon the existing work of \cite{he2022density}, which implements an autoencoder for LiDAR point cloud compression. Their method encodes point clouds while preserving local geometry and density by using three embeddings: \textit{density embedding}, \textit{local position embedding} and \textit{ancestor embedding}. It also allows for the compression of point-wise features, which is useful in our case for encoding relevant radar features like Doppler and RCS. An overview of our VAE model architecture is shown in Fig. \ref{fig:vae_architecture}. 

The encoder consists of several downsampling blocks, where at each stage the radar point cloud gets downsampled by a downsampling factor \(f_s\) using farthest point sampling. At each downsampling stage the collapsed points set \(C(p)\) is computed, which defines the discarded points that have collapsed into its nearest neighbor point \(p\) after downsampling. Each block then computes and aggregates the three separate feature embeddings. The \textit{density embedding} preserves the density of the point cloud by mapping the downsampling factor of each point, $u = |C(p)|$, to a $d$-dimensional representation using a multi-layer perceptron (MLP). The \textit{local position embedding} captures the distribution of $C(p)$ by encoding the local offset (direction and distance) of each collapsed point relative to its nearest neighbor. These local features are projected into a $d$-dimensional embedding through MLPs and self-attention layers. To pass along information from previous stages, a Point Transformer \cite{zhao2021point} layer is employed to aggregate features of the collapsed points set \(C(p)\) into the corresponding sampled point \(p\), producing the $d$-dimensional \textit{ancestor embedding}. Finally, an MLP combines all three embeddings into the feature representation for the next stage.

After the final downsampling step, the feature-enriched point cloud is projected into the latent space via an MLP. Sampling from this latent space gives us the latent point cloud representation. Before passing it to the decoder, we compute a more structured point cloud representation by obtaining the positional coordinates of each point using a PointNet \cite{qi2017pointnet} combined with features from an MLP.

The decoder follows a symmetric design relative to the encoder. At each decoding stage, the decoder block aims to reconstruct the original point cloud corresponding to the input of the matching encoding stage. The decoder block upsamples the downsampled point cloud by predicting an upsampling factor for each point, and then computing local offsets to obtain the upsampled points. After the final upsampling, the reconstructed radar point cloud is obtained, with point-wise features recovered via a segmentation head.

\subsection{Objective Function}
To train our VAE, we adopt the distortion loss from \cite{he2022density}, and add a Kullback-Leibler (KL) regularization term. The full objective function is thus defined as follows:

\begin{equation}
L_{\text{VAE}} = L_{rec} + \lambda_{den} L_{den} + \lambda_{card} L_{card} + \lambda_{reg} L_{reg}.
\end{equation}

\noindent
Here \(L_{rec}\) is the reconstruction loss, which encourages the model to accurately reconstruct both the point coordinates \(P \in \mathbb{R}^{N \times 3}\) and their associated point-wise features \(F \in \mathbb{R}^{N \times 2}\) of the original point cloud \(x \in \mathbb{R}^{N \times 5}\).
 We define this loss function as:

\begin{equation}
L_{rec} = L_{CD}(P, \tilde{P}) + \lambda_c \sum_{s=1}^{S} L_{CD}(P_s, \tilde{P}_s) + \lambda_f L_{feat}(F,\tilde{F}),
\end{equation}
\\
\noindent
where we use \(L_{CD}\), the Chamfer Distance (CD), to measure the positional difference between the original point cloud 3D coordinates \(P\) and the reconstructed point cloud coordinates \(\tilde{P}\). Given the symmetric design between our encoder and decoder, we also enforce that the intermediate coordinate reconstructions are similar using a Chamfer Distance over all \(S\) stages. The CD between two point cloud sets \(A\) and \(B\) is given by:

\begin{equation}
L_{CD}(A, B) = \frac{1}{|A|} \sum_{a \in A} \min_{b \in B} \| a - b \|_2^2 
+ \frac{1}{|B|} \sum_{b \in B} \min_{a \in A} \| a - b \|_2^2.
\end{equation}
\\
\noindent
Regarding feature reconstruction, we assign each point of the reconstructed point cloud to its nearest ground truth point. Afterwards, we simply compute an L2 loss between the predicted features \(\tilde{F}\) and their assigned ground truth features \(F\):
\begin{equation}
\label{eq:feat_loss}
L_{feat}(F,\tilde{F}) = \frac{1}{\tilde{N}} \sum_{i=1}^{\tilde{N}} \| f_{\pi(i)} - \tilde{f}_i \|_2^2,
\end{equation}

\noindent
where 
\(
\pi(i) = \arg\min_j \| p_j - \tilde{p}_i \|_2
\)
assigns each reconstructed point \(\tilde{p}_i\) to its nearest ground truth point \(p_j\).

The density loss \(L_{den}\) helps to recover the local density when upsampling. At each decoder stage \(s\), for each point \(\tilde{p}\) we want to predict the correct upsampling factor and the correct local offset to obtain the upsampled points, which is captured by:

\begin{equation}
     L_{den} = \sum_{s=1}^{S} \sum_{\tilde{p} \in \tilde{P}_{s}} \frac{\left||C(p) | - | \tilde{C}(\tilde{p}) |\right| +
     \lambda_d \left| \overline{C(p)} - \overline{\tilde{C}(\tilde{p})} \right|}{| \tilde{P}_{s} | },
\end{equation}
\\
\noindent
where the first term of the numerator measures the difference in the number of points between the downsampled points set \(C(p)\) and the upsampled points set \(\tilde{C}(\tilde{p})\). The second term computes the difference between the mean distances of all points in each set with respect to their center point \(p\) or \(\tilde{p}\).

Instead of only computing the density loss between the upsampled and downsampled points sets at each stage \(s\), we also incorporate a cardinality loss. This loss measures the difference in the number of points between the ground truth point cloud \(P_s\) and the reconstructed point cloud \(\tilde{P}_s\):

\begin{equation}
     L_{card} = \sum_{s=1}^{S} \left|\, |\mathcal{P}_s| - |\tilde{\mathcal{P}}_s| \,\right|.
\end{equation}

\noindent
Lastly, we regularize our latent point cloud space to follow a standard normal distribution using the Kullback-Leibler regularization:

\begin{equation}
L_{reg} = D_{\mathrm{KL}}\big(\mathcal{E}(z|x) \,\|\, p(z)\big).
\end{equation}

\section{Additional Implementation Details}\label{sup:implementation_details}
\textbf{View-of-Delft (VoD).} For VoD \cite{palffy2022multi}, we set our point cloud range to [0, 51.2], [-25.6, 25.6], [-3, 2] meters for the x, y, and z axes, respectively. Due to the sparsity of radar point clouds, we use five aggregated sweeps of radar point cloud data for all experiments on VoD in the same manner as \cite{palffy2022multi}. This includes training both our generative models and all object detectors. The bounding boxes of VoD do not contain any velocity information, which we require for the foreground generation. Therefore, we adopt a similar approach to nuScenes \cite{caesar2020nuscenes} for obtaining the velocities of the boxes.\\

\noindent
\textbf{TruckScenes.} For TruckScenes \cite{fent2024truckscenes}, we use a point cloud range of [-75, 75], [-75, 75] and [-2.5, 4.5] meters along the x, y, and z axes, respectively. Although the official evaluation includes detection ranges of up to 150 meters for certain classes, we restrict our range to half of this for simplicity and computational efficiency. Nevertheless, this setting still represents a substantially longer range than VoD and contains radar data with 360-degree coverage. In this case, we use six sweeps of radar point clouds following the official implementation of \cite{fent2024truckscenes}. Since TruckScenes does not provide compensated Doppler features for radar points, we perform ego-motion compensation ourselves to obtain these features. \\

\noindent
\textbf{Generative Models Training.} We use the following five features for radar point cloud generation: x, y, z, compensated Doppler, and RCS. The compensated Doppler is more intuitive and generally preferred over non-compensated Doppler, as it reflects the true radial velocity of an object by removing the influence of the ego vehicle’s motion. Consequently, it is easier for the LDM to generate accurate compensated Doppler features. 

When training our VAE and LDM, we normalize the radar point cloud data globally into [-1, 1] for all features. To support parallel training with fixed-size batches, we pad or randomly downsample each point cloud sample to match the number of points specified for training. For foreground generation specifically, the number of points in a scene often falls well below the specified fixed number of points. For example, when a scene only contains one or two bounding boxes with only a few points. In these cases, we augment the scene with additional bounding boxes and points from other samples using a technique similar to PolarMix \cite{xiao2022polarmix}. The hyperparameters for training our VAE and LDM are listed in Tabs. \ref{tab:vae_params} and \ref{tab:diff_params}, respectively. \\

\noindent
\textbf{Object Detection Training.} We use standard global augmentation techniques for all object detectors during training: random horizontal flip, global rotation within [\(-\pi/4\), \(\pi/4\)], and global scaling in the range of [0.95, 1.05]. Most of the annotations of VoD have an unknown difficulty label. Therefore, we do not filter annotations based on difficulty when using GT-Sampling. Moreover, we only insert bounding boxes that contain at least five radar points. 

The CenterPoint \cite{yin2021centerpoint} model is trained on VoD for 80 epochs using a batch size of 32. We use the AdamW optimizer with a weight decay of 0.01 and a one-cycle learning rate policy with a max learning rate of \(1\mathrm{e}{-3}\).

For TruckScenes dataset, all object detectors are trained for 30 epochs with a batch size of 4. We use the same optimizer and learning rate scheduler as VoD. 
\\

\noindent
\textbf{Evaluation Metrics.} 
We introduce a metric that captures the quality of the generated radar features, i.e., Doppler and RCS. The real radar point cloud is denoted as \(P_{\mathrm{gt}} = \{ u_1, u_2, \ldots, u_n\},\) and the generated radar point cloud as \(
P_{\mathrm{gen}} = \{ v_1, v_2, \ldots, v_m\}, \) where each point \(u_i\) and \(v_j\) consists of a position vector (x, y, z) and feature vector composed of Doppler and RCS values. For each generated point \(v_j \in P_{\mathrm{gen}}\), we find its nearest neighbor in the ground truth point cloud:

\begin{equation}
u^*_j = \underset{u_i \in P_{\mathrm{gt}}}{\arg\min} \| p(v_j) - p(u_i) \|_2,
\end{equation}

\noindent
where \(p(v_j)\) and \(p(u_i)\) denote the 3D positions of \(v_j\) and \(u_i\), respectively.  The generation quality for a single feature \(f\) (Doppler or RCS) is then defined as:

\begin{equation}
\text{CD}_f = \frac{1}{m} \sum_{j=1}^{m} \big| f(v_j) - f(u^*_j) \big|.
\end{equation}

\begin{table}[h!]
\centering
\caption{Hyperparameters for VAE training. Batch size differs based on foreground / background generation task.}
\label{tab:vae_params}
\small
\begin{tabular}{lc} 
\toprule
Hyperparameter & Value \\ 
\midrule
Number of epochs & 300 \\
Batch size & 128 / 32 \\
Learning rate & \(1\mathrm{e}{-3}\) \\
Optimizer & Adam \\
Scheduler & StepLR \\
step size of StepLR & 45 \\
\(\gamma\) of StepLR & 0.5 \\
KL weight (\(\lambda_{reg}\)) & \(1\mathrm{e}{-5}\) \\
Density weight (\(\lambda_{den}\)) & \(1\mathrm{e}{-4}\) \\
Cardinality weight (\(\lambda_{card}\)) & \(5\mathrm{e}{-7}\) \\
Mean distance weight (\(\lambda_d\)) & 50.0 \\
Intermediate chamfer weight (\(\lambda_{c}\)) & 0.1 \\
Feature weight (\(\lambda_{f}\)) & 0.05 \\
Latent point dimension (\(d_z\)) & 4 \\
\bottomrule
\end{tabular}
\end{table}

\begin{table}[h!]
\centering
\caption{Hyperparameters for latent diffusion model training. Batch size differs based on foreground / background generation task.}
\label{tab:diff_params}
\small
\begin{tabular}{lc} 
\toprule
Hyperparameter & Value \\ 
\midrule
Number of epochs & 1000 \\
Batch size & 128 / 16 \\
Learning rate & \(1\mathrm{e}{-4}\) \\
Optimizer & AdamW \\
weight decay of AdamW & \(1\mathrm{e}{-6}\) \\
Scheduler & OneCycleLR \\
\midrule
\multicolumn{2}{c}{Diffusion Parameters} \\
\midrule 
\(\beta_0\) & \(1\mathrm{e}{-4}\) \\
\(\beta_T\) & 0.02 \\
\(\beta_t\) schedule & Linear \\
Diffusion Steps \(T\) & 1000 \\
\bottomrule
\end{tabular}
\end{table}

\section{Ablation Study on Backbone}\label{sup:ablation}
We investigate different choices of backbone for our denoising network. Specifically, we consider several architectures used in the point cloud processing or point cloud diffusion literature: PointTransformer (PT) \cite{zhao2021point}, Point Transformer V3 (PTv3) \cite{wu2024pointv3}, 3D Diffusion Transformers (DiT-3D) \cite{mo2023dit3d}, Point-Voxel CNN (PVCNN) \cite{liu2019pvcnn}, and Sparse Point-Voxel Diffusion (SPVD) \cite{romanelis2024efficient}. We evaluate the effectiveness of each backbone based on the generation quality of radar point clouds for the task of unconditional generation. We assess the quality using quantitative metrics such as Jensen-Shannon divergence (JSD) and Minimum Matching Distance (MMD) adapted from LiDAR-based diffusion \cite{004ran2024LiDMs}.  

\begin{table}[!htb]\centering
    \caption{Ablation on backbone of the latent diffusion model. Comparison of unconditional radar point cloud generation on VoD dataset \cite{palffy2022multi}. MMD is multiplied by $1 \times 10^{4}$.}
    \label{tab:ablation_backbone}
    \small
    \begin{tabular}{lcc}
        \toprule
       Method & JSD~$\downarrow$ &  MMD~$\downarrow$ \\
        \midrule
        PT \cite{zhao2021point} & 0.252 & 6.92 \\
        PTv3 \cite{wu2024pointv3} & 0.243 & 6.66 \\
        DiT-3D \cite{mo2023dit3d} & 0.245 & 6.99  \\
        PVCNN \cite{liu2019pvcnn} & 0.238 & 6.41 \\
        \textbf{SPVD} \cite{romanelis2024efficient} & \textbf{0.233} & \textbf{6.25}  \\
        \bottomrule
    \end{tabular}
\end{table}

\begin{figure*}[tb] \centering
\includegraphics[width=0.98\linewidth, height=0.98\linewidth]{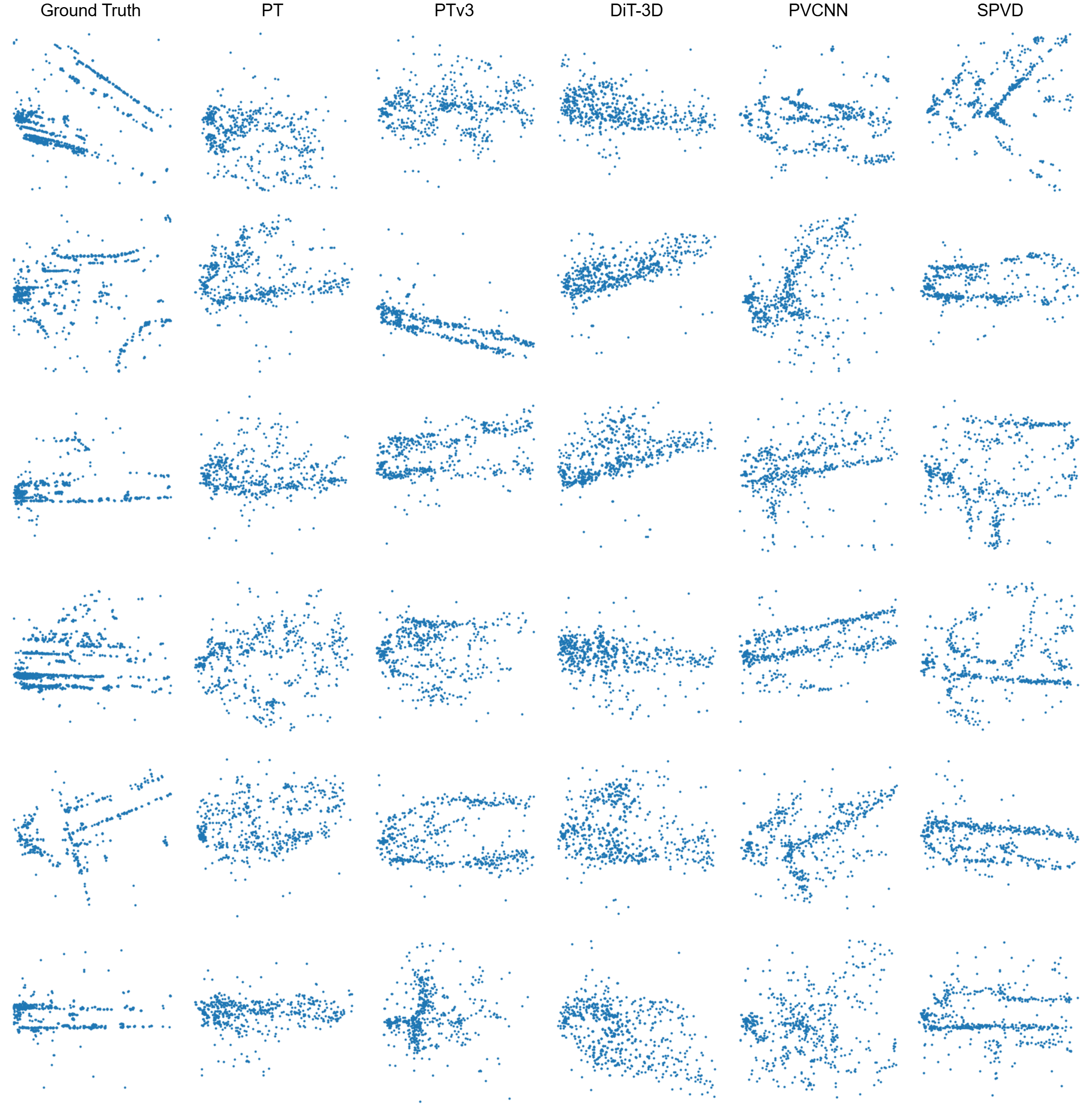}
    \caption{Qualitative results on the VoD dataset \cite{palffy2022multi} of \textbf{unconditional} radar point cloud generation using various backbones for the latent diffusion model. All models use the same VAE.} 
\label{fig:suppl_backbone_comparison}
\end{figure*}

From Tab.~\ref{tab:ablation_backbone}, SPVD achieves slightly better performance than the other methods. However, we note that these quantitative metrics may not be ideal for evaluating unconditional generation of radar point clouds, as they consider only spatial similarity and do not account for radar-specific features. Furthermore, we believe the small differences in the metrics can be attributed to the inherent sparsity of radar point cloud data and the fact that radar points are very noisy. Therefore, we also provide a qualitative comparison of unconditional radar point cloud generation of the various models in Fig. \ref{fig:suppl_backbone_comparison}. We observe that SPVD appears to generate more accurate and diverse representations of radar point clouds compared to the other approaches. 

\section{Additional Qualitative Results}\label{sup:add_results}
\begin{figure*}[tb] \centering
\includegraphics[width=0.98\linewidth, height=0.7546\linewidth]{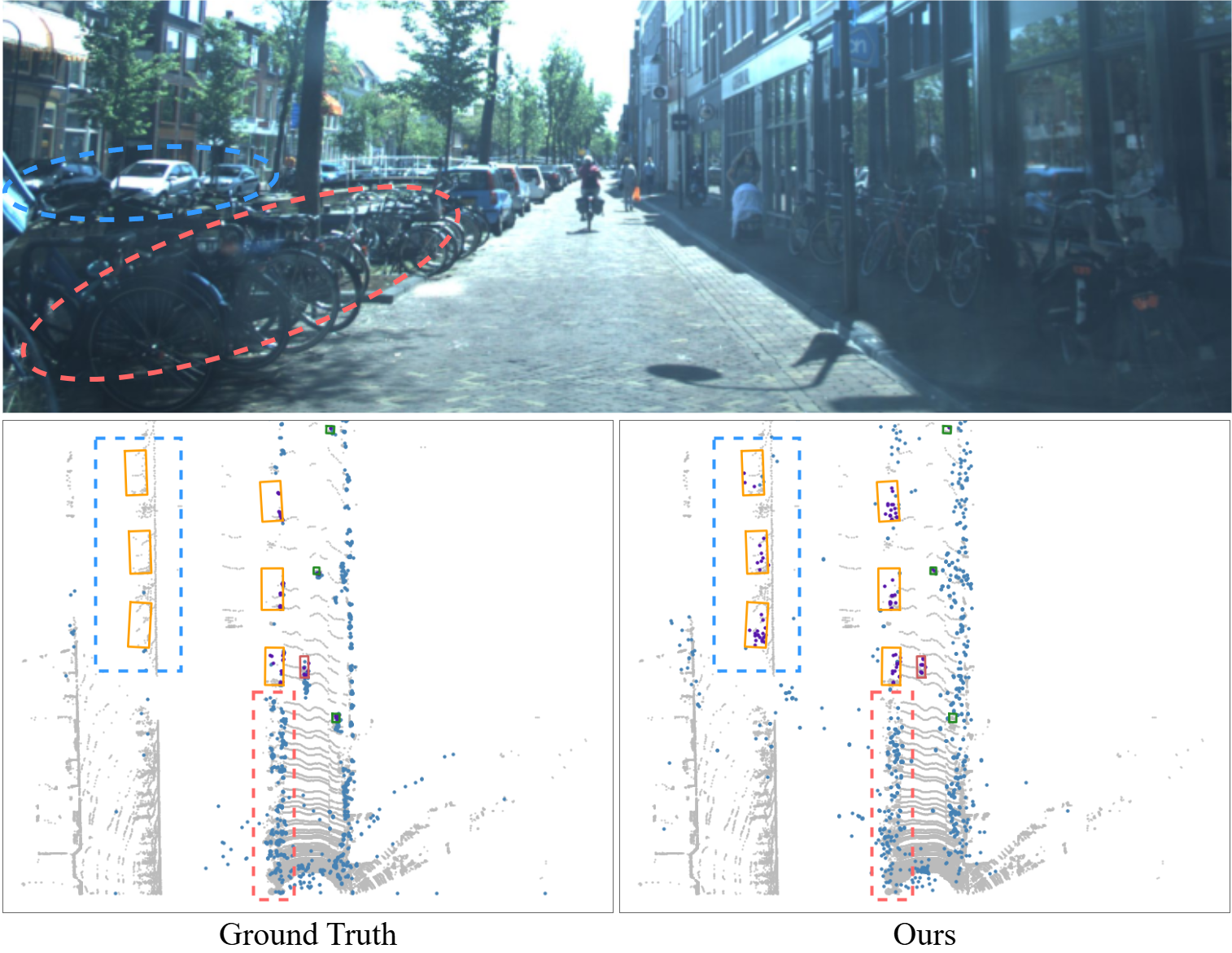}
    \caption{Limitation of our method on VoD \cite{palffy2022multi}. Due to the separation of foreground and background generation, we still generate foreground points for objects (\textcolor[RGB]{51,153,255}{dashed blue}) which are partially occluded by the environment (\textcolor[RGB]{255,102,102}{dashed red}).}
    \label{fig:suppl_failure}
\end{figure*}
We present a limitation of our approach in Fig.~\ref{fig:suppl_failure}. Since we separate the generation process for background and foreground points, the two components are generated independently. As a result, for bounding boxes that are partially occluded by the environment (background points), our model may still generate foreground points even though the ground truth contains none.

\begin{figure*}[tb] \centering
\includegraphics[width=\textwidth, height=1.216\textwidth]{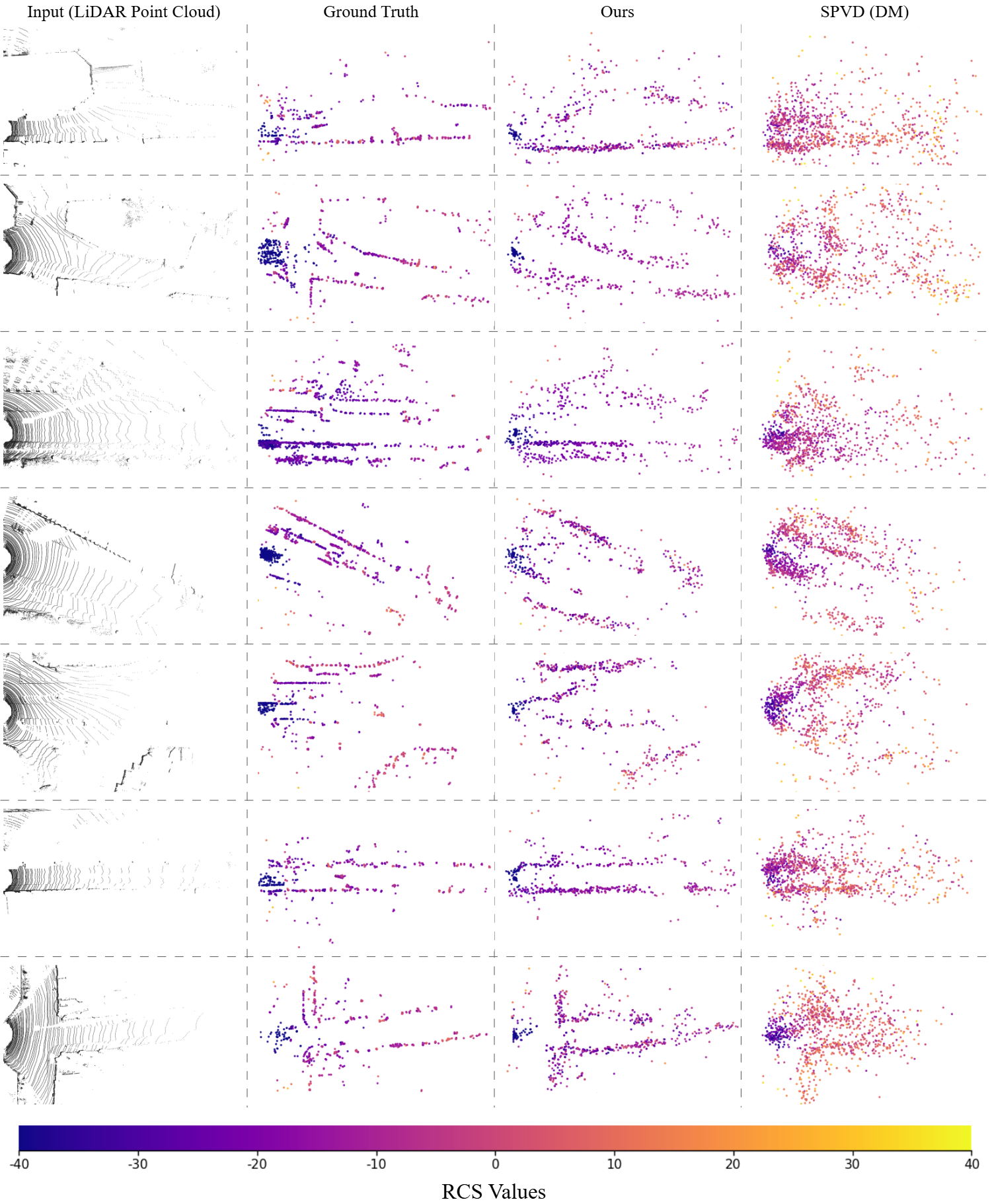}
    \caption{Qualitative results on the VoD dataset \cite{palffy2022multi} of background radar point cloud generation based on LiDAR as conditioning mechanism, radar points are colored by RCS values.} 
\label{fig:suppl_qualitative_background}
\end{figure*}

We also provide qualitative results of background generation on the VoD dataset in Fig. \ref{fig:suppl_qualitative_background}. In this figure, we compare background points generated by our method against those from the vanilla diffusion model SPVD (DM) used in the ablation study (Sec.~\ref{subsec:ablation}). We can see that the diffusion model captures the overall scene structure, but it produces highly noisy points with unrealistic RCS distributions. In contrast, our method better preserves the fine-grained details of radar point clouds while maintaining realistic RCS values for the radar points.

\begin{figure*}[tb] \centering
\includegraphics[width=\textwidth, height=1.09\textwidth]{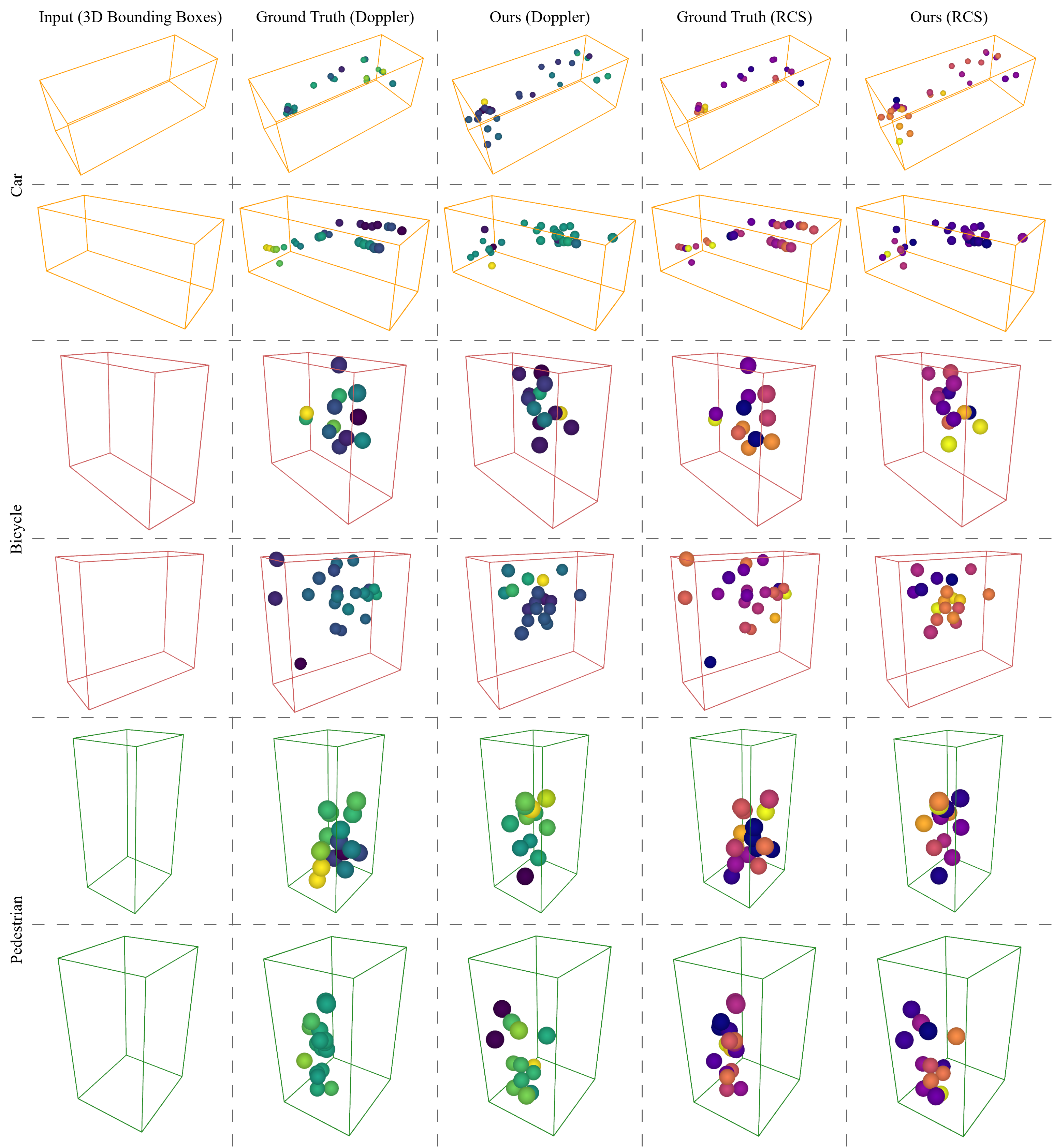}
    \caption{Qualitative results on the TruckScenes dataset \cite{fent2024truckscenes} of foreground radar point generation based on 3D bounding boxes as conditioning mechanism. Note that for conditioning we actually provide multiple bounding boxes of a scene, but the visualization shows only a single bounding box for clarity. Points are colored by Doppler or RCS (color scale normalized per bounding box).} 
\label{fig:suppl_qualitative_foreground}
\end{figure*}

Moreover, qualitative results of foreground generation on the TruckScenes dataset can be seen in Fig. \ref{fig:suppl_qualitative_foreground}. Our approach generates not only accurate 3D positions for foreground points but also reliable Doppler and RCS values across multiple classes.

\end{document}